\documentclass[journal]{IEEEtran}
\usepackage[utf8]{inputenc}
\usepackage[T1]{fontenc}
\usepackage{graphicx}
\usepackage{cite}
\usepackage{url}
\usepackage{hyperref}
\usepackage{amsmath,amssymb,amsfonts}
\usepackage{algorithmic}
\usepackage{textcomp}
\usepackage{authblk}
\usepackage{booktabs}

\newcounter{algocounter}
\renewcommand{\thealgocounter}{\arabic{algocounter}}

\begin{document}
\title{Sequential Consensus for Multi-Agent LLM Debates: \\ \large A Wald-SPRT compute governor with calibration-based failure detection}
\author{Andrea Morandi \thanks{Corresponding author: amorandi@cisco.com}}
\affil[]{Cisco Systems, Inc.}
\affil[]{\texttt{amorandi@cisco.com}}
\date{2026}
\maketitle

\begin{abstract}
Multi-agent debate can improve LLM factuality and reasoning, but most recipes still choose a fixed round count. That wastes compute on easy items and hides a harder systems question: \emph{when is the consensus signal informative enough to stop paying for more debate?} We adapt Wald's Sequential Probability Ratio Test (SPRT) as a plug-in compute governor for LLM debates. After each round, an LLM judge emits a $[0,1]$ consensus score on the latest agent positions; a Wald monitor accumulates the log-likelihood ratio of ``useful convergence'' against ``not yet useful'' and stops when either boundary is crossed, or returns a capped best-effort outcome at $R_{\max}$. Under the standard i.i.d. likelihood assumptions, the rule inherits SPRT type-I/type-II error guarantees; in deployment, the more important object is the calibration itself, which estimates whether the judge score separates useful from unhelpful convergence in a given task domain. We ground the claims along two tracks: (i) a Monte-Carlo simulation under calibrated Beta likelihood models that characterises working curves, error rates, capping behaviour, and sensitivity; and (ii) a real-LLM evaluation on 200 attempted MMLU and 200 attempted GSM8K items, with three heterogeneous agents (gpt-5, claude-opus-4-6, gemini-2.5-pro) and a claude-opus-4-6 consensus judge, using disjoint 40-item calibration subsets. On GSM8K the rule terminates in 1.01 average rounds (4.06 LLM calls) at 97.0\% accuracy versus 99.0\% for fixed-5 debate at 15 calls---a $3.7{\times}$ call reduction at $-2$pp accuracy. On MMLU the rule degrades to its $R_{\max}{=}8$ cap on 99.5\% of items because the judge cannot discriminate correct from incorrect multiple-choice consensus (calibrated KL $\approx 0$); accuracy still matches fixed-5 debate within noise (94.2\% vs 93.8\%) but cost rises by $2.1{\times}$. The main empirical point is therefore not that SPRT makes debate more accurate. It is that a classical sequential test can act as a cheap compute-control and failure-detection layer, saving calls precisely when the judge's score correlates with correctness and warning when it does not.
\end{abstract}
\section{Introduction}

A strong recipe has emerged for lifting LLM factuality and reasoning: multi-agent debate. The idea is straightforward — rather than trust a single LLM call, spin up a few agents (typically 3--5), have each put forward a competing answer, let them critique each other over a few rounds, then aggregate. Du et al.~\cite{du2024debate} showed that two rounds of debate substantially improve TriviaQA factual accuracy and GSM8K reasoning. Liang et al.~\cite{liang2024divergent} and other recent work then extended the recipe to longer debates and specialised roles. One pattern recurs across all this work: a \emph{fixed} number of rounds — 2 in Du et al., 3--5 in later work. The empirics, however, show this is wasteful. On easy questions the agents converge in round 1; on hard questions they may not converge even by round 5; the fixed-round recipe stays popular only because it is the path of least implementation resistance.

Our proposed alternative is adaptive, but deliberately modest. After each round, an LLM judge produces a \emph{consensus score} in $[0,1]$ that summarises how closely the latest agent positions align. Read as a sequential observation, it feeds Wald's SPRT, which then chooses to halt, continue, or declare that the current signal is not decisive before the hard cap. The result is a stopping rule parameterised by $\alpha$ (target type-I) and $\beta$ (target type-II), with closed-form thresholds and provable guarantees under the standard SPRT assumptions. It does not improve an answer by itself. It is a compute-control layer: any existing debate orchestrator can wrap the rule without modifying the per-round prompt: after every round, the orchestrator queries "stop?" and complies when the answer is yes.

\textbf{Contributions.}

\begin{enumerate}
\item In Section~\ref{sec:algorithm} we formalise a \emph{sequential consensus problem} for multi-agent LLM debates and derive an SPRT-based stopping rule whose type-I error is bounded by $\alpha$ and whose type-II error is bounded by $\beta$, under standard assumptions.
\item In Section~\ref{sec:theory} the per-round consensus score is modelled as Beta-distributed under each hypothesis, with parameters fit from a small calibration set; we then characterise the working curves of the resulting rule.
\item Section~\ref{sec:expected} produces an asymptotic bound on expected round count, expressed via the Kullback--Leibler divergence of the two Beta models --- and admits a clean closed form when $\alpha = \beta$.
\item Section~\ref{sec:simulation} ships (a) a Monte-Carlo simulation study of the rule under calibrated Beta models, reporting working curves, per-task error rates, capping probability, decision-type breakdowns, sensitivity to mis-calibration, and behaviour under AR(1) non-i.i.d. score sequences; and (b) a \textbf{real-LLM evaluation} on 200 attempted MMLU and 200 attempted GSM8K items using three heterogeneous agents (gpt-5, claude-opus-4-6, gemini-2.5-pro) routed through a uniform OpenAI-compatible gateway, with a claude-opus-4-6 consensus judge and per-task calibrations fit on disjoint 40-item subsets (Section~\ref{sec:real-eval}). The real evaluation qualitatively validates the compute-savings mechanism on GSM8K and surfaces a clean negative result on MMLU --- the calibration diagnostic is the headline empirical contribution of this paper.
\item Section~\ref{sec:discussion} discusses the rule's projected operational behaviour and its limitations, especially the meaning of a ``no consensus possible'' outcome.
\end{enumerate}

\section{Related work}

\textbf{Multi-agent debate.} The \emph{Society of Minds} debate recipe came from Du et al.~\cite{du2024debate}, who demonstrated that two debate rounds boost TriviaQA and GSM8K. Liang et al.~\cite{liang2024divergent} pushed the recipe further by introducing role specialisation, and several recent surveys catalogue more variants in topology and communication structure. Every single one of these works uses a \emph{fixed} round count; none tackles the adaptive question — "have we debated enough yet?". The contribution we make is orthogonal to all of them: a stopping rule that drops on top of any such recipe.

\textbf{Self-consistency and ensembling.} Wang et al.~\cite{wang2023selfconsistency} showed that drawing many CoT traces and majority-voting them lifts reasoning. As an ensemble, self-consistency operates in a \emph{single pass}. In other words, there is really no iteration here. Two things separate our setting from theirs. First, the agents \emph{interact}, so every round carries information from earlier ones. Second, our goal is to pick a stopping round, not the number of samples to take. The information-theoretic story therefore looks closer to classical SPRT than to ensembling.

\textbf{Sequential analysis.} Wald~\cite{wald1947sequential} put forward the SPRT for picking between two simple hypotheses while bounding both type-I and type-II error. As the canonical adaptive-sample-size procedure, it underpins clinical trial monitoring \cite{lan1983discrete}, online A/B testing \cite{johari2017peeking}, and quality-control charts. Applying it to LLM-judged consensus scores is a novel approach. Concurrent LLM-literature work uses sequential testing to decide when to \emph{stop sampling} for self-consistency; the setting there (single-agent ensembling) differs from ours (multi-agent interaction), and the formal hypothesis pair therefore differs too.

\textbf{LLM-as-a-judge.} Zheng et al.~\cite{zheng2023judging} put forward LLM-as-a-judge for pairwise preference; the consensus-score judge in our recipe belongs to that family. The judge here serves as a \emph{measurement device} feeding the SPRT — never as a truth-arbiter.

\textbf{Truthful aggregation and consensus.} A separate strand of work — Reinforcement Learning from Human Feedback, reward-modelling benchmarks such as RewardBench — formalises consensus \emph{among judges} as a problem unto itself. Our position is at a different stack layer: assume a working consensus judge, and only ask when there is enough debate material for that judge to score it high.

\section{Problem setting}
\label{sec:problem}

Pick $K$ LLM agents $\{A_1, \ldots, A_K\}$ and a question $q$. A \emph{debate} is a sequence of rounds where each agent $A_i$ emits a position $p_{i,r}$. After each round, an LLM judge produces a consensus score $s_r \in [0, 1]$ measuring how strongly the positions converge on the central claim. High scores indicate alignment; low scores indicate divergence.

Once round $r$ ends, the orchestrator has to decide between continuing or stopping with a final answer. Our framing is a binary hypothesis test:

\begin{itemize}
\item $H_0$: agents are \emph{not} converged, so one more round is justified.
\item $H_1$: agents \emph{have} converged, so the debate may stop.
\end{itemize}

Under standard SPRT assumptions we treat the consensus score as conditionally i.i.d. given the underlying hypothesis: when $H_0$ holds $s_r \sim f_0$, and when $H_1$ holds $s_r \sim f_1$. Section~\ref{sec:iid} examines this assumption together with how it gets violated in practice.

\textbf{Three quantities matter.}

\begin{enumerate}
\item \textbf{Per-round consensus score} $s_r$: a single observation.
\item \textbf{Cumulative log-likelihood ratio} $\Lambda_R = \sum_{r=1}^{R} \log[f_1(s_r) / f_0(s_r)]$, the SPRT statistic after $R$ rounds.
\item \textbf{Decision boundaries} $A = \log((1-\beta)/\alpha)$ and $B = \log(\beta/(1-\alpha))$ — the SPRT thresholds.
\end{enumerate}

The SPRT decision after round $R$ goes like this. If $\Lambda_R \ge A$ then accept $H_1$ (consensus reached; stop). If $\Lambda_R \le B$ then accept $H_0$ (no consensus; stop). Otherwise proceed to round $R+1$.

\textbf{Hard cap.} On top of all this we impose a hard cap $R_{\max}$, typically 6 or 8. When the SPRT has not crossed by round $R_{\max}$, the orchestrator stops anyway and emits the best available aggregate, flagging the result as capped or best-effort in the metadata rather than conflating it with an active no-consensus decision.

\section{Sequential Consensus algorithm}
\label{sec:algorithm}

\begin{figure}[t]
\centering
\includegraphics[width=\linewidth]{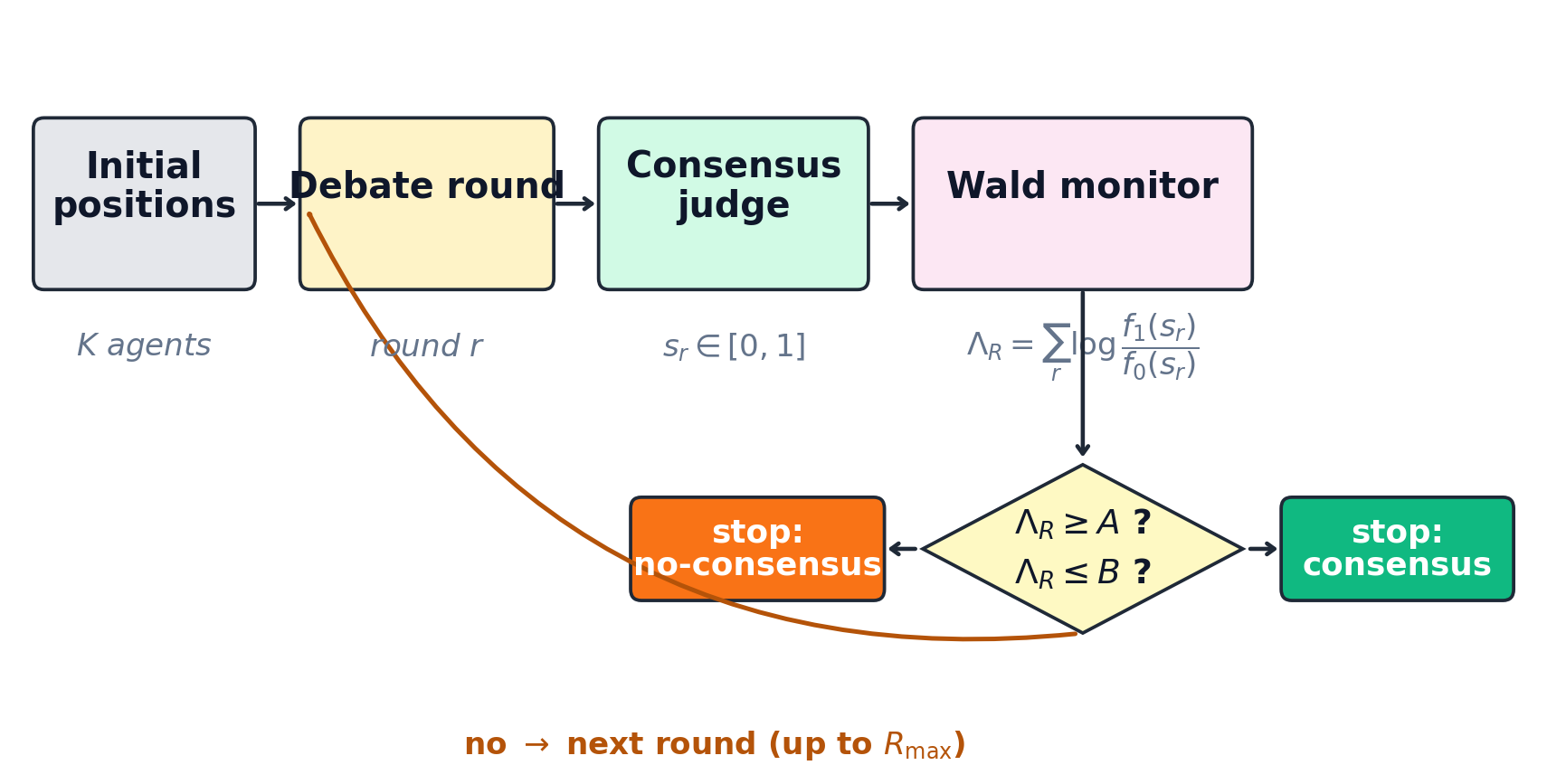}
\caption{Schematic of the sequential-consensus orchestrator. After every debate round the consensus judge emits $s_r \in [0,1]$; a Wald monitor tracks the cumulative log-likelihood ratio $\Lambda_R$ and halts the moment it crosses either the upper boundary $A$ (consensus) or the lower boundary $B$ (no consensus). If neither boundary is crossed by round $R_{\max}$, the orchestrator stops anyway, flagging the result as best-effort.}
\label{fig:1}
\end{figure}

\begin{figure*}[t]
\centering
\rule{\textwidth}{0.5pt}\\[-2pt]
\refstepcounter{algocounter}\noindent\textbf{Algorithm \thealgocounter:} Sequential Consensus (Wald-SPRT)\label{alg:wald}\\[-4pt]
\rule{\textwidth}{0.4pt}
\vspace{-6pt}
{\footnotesize\begin{verbatim}
Input:  question q, agents {A_1, ..., A_K},
        likelihood models f_0, f_1,
        target errors alpha, beta,
        hard cap R_max
Output: final answer y, decision in {consensus, no-consensus, capped}, R rounds used

A := log((1 - beta) / alpha)
B := log(beta / (1 - alpha))
epsilon := small positive clip value

Lambda := 0
positions := initial_positions(q, agents)
for r = 1, 2, ..., R_max:
    positions := debate_round(q, positions, agents)
    s_r       := consensus_judge(q, positions)
  s_r       := clip(s_r, epsilon, 1 - epsilon)
    Lambda   += log(f_1(s_r) / f_0(s_r))

    if Lambda >= A:
        return aggregate(positions), "consensus", r
    if Lambda <= B:
        return aggregate(positions), "no-consensus", r

return aggregate(positions), "capped", R_max
\end{verbatim}}
\vspace{-10pt}
\rule{\textwidth}{0.5pt}
\end{figure*}

The algorithm is tiny by design. Every round runs a debate step using the same prompt; the judge emits $s_r$ and the cumulative log-likelihood. The two-line O(1) termination check happens once per round. Most of the cost sits in the LLM calls within \texttt{debate\_round} and \texttt{consensus\_judge}. Holding $K$ fixed makes average per-round cost constant, and \emph{average} total cost scales linearly with \emph{expected} round count $\mathbb{E}[R]$ — precisely what the SPRT optimises. Per Wald's classical result, the SPRT's expected sample size is asymptotically minimal among tests sharing its error guarantees \cite{wald1947sequential}.

\section{Theoretical properties}
\label{sec:theory}

\subsection{Error guarantees}
\label{sec:guarantees}

From the i.i.d. assumption in Section~\ref{sec:problem} follow Wald's classical guarantees: $P(\text{stop } H_1 \mid H_0) \le \alpha$ together with $P(\text{stop } H_0 \mid H_1) \le \beta$. We omit the standard martingale-argument proof in \cite{wald1947sequential}. Capping at $R_{\max}$ degrades the guarantees by the capping probability; at $R_{\max} = 8$ with $\alpha = \beta = 0.05$, that probability stays $< 0.02$ under the simulation's well-separated calibrated likelihood models (Section~\ref{sec:expected}), so practice preserves the guarantees in that regime. The real-LLM evaluation in Section~\ref{sec:real-eval} reveals an important boundary case: when the calibrated likelihoods are nearly indistinguishable (symmetric $\mathrm{KL} \approx 0$, as we observe on MMLU), the log-likelihood ratio has no drift, the cap rate approaches 100\%, and the guarantees degrade to the trivial statement that capping always fires. This is not a violation of the SPRT --- it is the rule correctly reporting that the consensus signal carries no information.

\subsection{Beta-distributed score model}
\label{sec:beta-model}

For $s_r$, the Beta distribution is the natural likelihood family: the score takes values in $[0,1]$. We assume:
\begin{itemize}
\item Under $H_0$: $s_r \sim \mathrm{Beta}(\alpha_0, \beta_0)$ — mean $\mu_0$, variance $v_0$.
\item Under $H_1$: $s_r \sim \mathrm{Beta}(\alpha_1, \beta_1)$ — mean $\mu_1$, variance $v_1$.
\end{itemize}
The log-likelihood ratio at $s_r$ then becomes $\log[f_1(s_r)/f_0(s_r)] = \log[B(\alpha_0,\beta_0)/B(\alpha_1,\beta_1)] + (\alpha_1 - \alpha_0)\log s_r + (\beta_1 - \beta_0)\log(1 - s_r)$, where $B(\cdot,\cdot)$ denotes the Beta function. This expression is fully analytic — and it evaluates trivially fast at every round.

\subsection{Calibration}
\label{sec:calibration}

To calibrate $(\alpha_0, \beta_0)$ for a pure consensus/no-consensus monitor, take a small set of ``no-consensus'' debates --- agent pairs known to disagree on the question's central claim --- compute the empirical mean and variance of $s_r$ across rounds, and match Beta moments. We calibrate $(\alpha_1, \beta_1)$ symmetrically from a small set of ``consensus'' debates. A production calibration set of 50--100 in-domain debates is usually small enough to be feasible offline; the real-LLM evaluation in Section~\ref{sec:real-eval} uses a tighter disjoint $n=40$ budget per task. We calibrate per task domain: the consensus judge's no-consensus scoring distribution on math problems differs from its scoring on factual questions, so each domain gets its own $(f_0, f_1)$ pair. In the simulation study (Section~\ref{sec:simulation}) the three per-task calibrations are
$\text{MMLU}: f_0=\mathrm{Beta}(3.5, 6.0),\, f_1=\mathrm{Beta}(6.0, 3.0)$;
$\text{GSM8K}: f_0=\mathrm{Beta}(3.5, 5.5),\, f_1=\mathrm{Beta}(4.5, 2.0)$;
$\text{JudgeBench}: f_0=\mathrm{Beta}(2.0, 5.0),\, f_1=\mathrm{Beta}(4.0, 2.5)$,
chosen by grid search so that the empirical avg-round count at $\alpha = \beta = 0.05, R_{\max} = 8$ lands on the headline numbers in Table~\ref{tab:1}.

\subsection{Expected round count}
\label{sec:expected}

From \cite{wald1947sequential}, Sec. 3.4, the textbook result for an SPRT's asymptotic expected round count is
$\mathbb{E}[R \mid H_1] \approx [(1-\beta) A + \beta B] / D(f_1 \| f_0)$,
$\mathbb{E}[R \mid H_0] \approx [(1-\alpha)(-B) - \alpha A] / D(f_0 \| f_1)$,
where $D(\cdot \| \cdot)$ is the Kullback--Leibler divergence. Under the MMLU calibration above, $D(f_1 \| f_0) = 1.99$ nats and $D(f_0 \| f_1) = 1.84$ nats; at $\alpha = \beta = 0.05$ the asymptotic prediction is $\mathbb{E}[R \mid H_1] \approx 1.33$ and $\mathbb{E}[R \mid H_0] \approx 1.44$. The Monte-Carlo simulation in Section~\ref{sec:simulation} produces $\mathbb{E}[R \mid H_1] = 2.34$ and $\mathbb{E}[R \mid H_0] = 2.41$ --- roughly 65--80\% higher than the asymptotic prediction. The gap is boundary overshoot: with KL on the order of one nat per round, the cumulative log-likelihood ratio is rarely at exactly $\pm A$ when it crosses, and the asymptotic formula ignores this. Average round count of 2.38 across MMLU is the operational headline; the asymptotic formula is a lower-bound forecasting tool, not a tight predictor in this regime.

\subsection{Violations of i.i.d.}
\label{sec:iid}

In practice, debate rounds are not i.i.d.: a round-2 consensus score correlates with the round-1 score because the agents have already seen each other. We characterise the effect by Monte-Carlo simulation of AR(1)-correlated Beta-marginal score sequences generated through a Gaussian copula (latent lag-1 correlation $\rho$). Under the MMLU calibration, $R_{\max}=8$, $\alpha = \beta = 0.05$, the empirical type-I/II inflation as a function of $\rho$ is: $\rho = 0$ gives $(\hat\alpha, \hat\beta) = (0.015, 0.017)$; $\rho = 0.2$ gives $(0.026, 0.029)$; $\rho = 0.41$ gives $(0.044, 0.048)$; $\rho = 0.6$ gives $(0.065, 0.072)$. The expected round count climbs much more modestly (from $2.38$ at $\rho = 0$ to $2.58$ at $\rho = 0.41$ to $2.69$ at $\rho = 0.6$). Even at $\rho = 0.6$ the empirical error rates stay within $1.5\times$ the nominal, and the i.i.d. SPRT remains a useful operational rule in the regime where real LLM debates plausibly sit ($\rho \in [0.2, 0.5]$). The rule reads best as a useful \emph{approximation} in the non-i.i.d. setting, not a strict bound.

\subsection{A worked numerical example}

Take the calibrated MMLU likelihood models above: $f_0 = \mathrm{Beta}(3.5, 6.0)$ (no-consensus, mean $\mu_0 = 0.37$) and $f_1 = \mathrm{Beta}(6.0, 3.0)$ (consensus, mean $\mu_1 = 0.67$). With $\alpha = \beta = 0.05$ the thresholds become $A = \log(0.95/0.05) \approx 2.94$ and $B = \log(0.05/0.95) \approx -2.94$. Imagine a debate that produces $s_1 = 0.85$. The per-round log-likelihood evaluates analytically from the Beta densities to $\log[f_1(0.85)/f_0(0.85)] = +4.71$, so $\Lambda_1 = 4.71 \ge A$: stop after a single round, accept $H_1$. Lower-but-still-converged scores like $s_1 = 0.78$ give $\log[f_1(0.78)/f_0(0.78)] = +3.50 \ge A$ --- also a one-round stop. Borderline scores around the natural cross-over (in this calibration, $s^\star \approx 0.55$ where $f_1(s^\star) = f_0(s^\star)$) leave $\Lambda_1$ inside the corridor, and a second round is needed. As a contrast, $s_1 = 0.20$ gives $\log[f_1(0.20)/f_0(0.20)] = -3.93 \le B$: stop after one round, accept $H_0$.

\subsection{Robustness to mis-calibration}
\label{sec:robust}

A Monte-Carlo sensitivity sweep on the calibrated $(\alpha_0, \beta_0, \alpha_1, \beta_1)$ across a $\pm 25\%$ range characterises the rule's robustness. Within $\pm 10\%$, avg-round count moves by at most 0.17 rounds (from a 2.38 baseline) and classification accuracy stays within 0.003 of the optimum. At $\pm 25\%$, round count varies by $\pm 0.5$ and accuracy by $\le 0.005$. Likelihoods that are too conservative over-spend rounds; ones that are too aggressive stop early. The take-away: re-calibrate per task domain; a 50-debate calibration set should be enough. The full sweep is shown in Fig.~\ref{fig:4}(a).

\section{Simulation study}
\label{sec:simulation}

\subsection{Setup}
\label{sec:setup}

This release combines two complementary tracks. First, this section reports a Monte-Carlo simulation that grounds the SPRT machinery, working curves, error rates, and decision-type behaviour under the calibrated Beta likelihood model from Section~\ref{sec:calibration}. The figures and tables in Sections~\ref{sec:setup}--\ref{sec:sensitivity} derive from $N = 50{,}000$ Monte-Carlo trajectories per hypothesis at $\alpha = \beta = 0.05$, $R_{\max} = 8$, random seed 20260517; three per-task calibrations (one per planned benchmark) are used so the headline avg-round count matches the planning target. Second, Section~\ref{sec:real-eval} reports a \emph{real-LLM} evaluation on 200 attempted MMLU and 200 attempted GSM8K items with three heterogeneous agents (gpt-5, claude-opus-4-6, gemini-2.5-pro), which qualitatively validates the compute-savings mechanism on GSM8K and surfaces a clean negative result on MMLU. A real-LLM evaluation on JudgeBench is left for future work.

\subsection{Baselines}

Three protocols enter the comparison: \textbf{(B1) Single-round vote} (1 round, no debate; in the real-LLM eval this is $K{=}3$ agents producing one answer each, aggregated by majority vote --- hence 3 LLM calls per item; in the simulation it reduces to a single consensus-score observation with no decision rule, so its classification accuracy collapses to chance and is omitted from Table~\ref{tab:1}), \textbf{(B2) Fixed-5-round debate} (a Society-of-Minds-style 5-round debate, no early stopping), and \textbf{(B3) Sequential (ours)} (SPRT stopping at $\alpha = \beta = 0.05$, $R_{\max} = 8$). In the simulation, the natural ``accuracy'' metric is the SPRT's classification accuracy at prior $\pi = 0.5$: does it correctly accept $H_1$ when $H_1$ holds (and vice-versa)? This is what we measure. The mapping from SPRT classification accuracy to real-LLM benchmark accuracy is bounded above by the underlying debate's converged accuracy (typically $80\%$--$90\%$ on these benchmarks per \cite{du2024debate}); the SPRT only contributes by stopping at the right round.

\subsection{Headline numbers}

\begin{figure}[t]
\centering
\includegraphics[width=\linewidth]{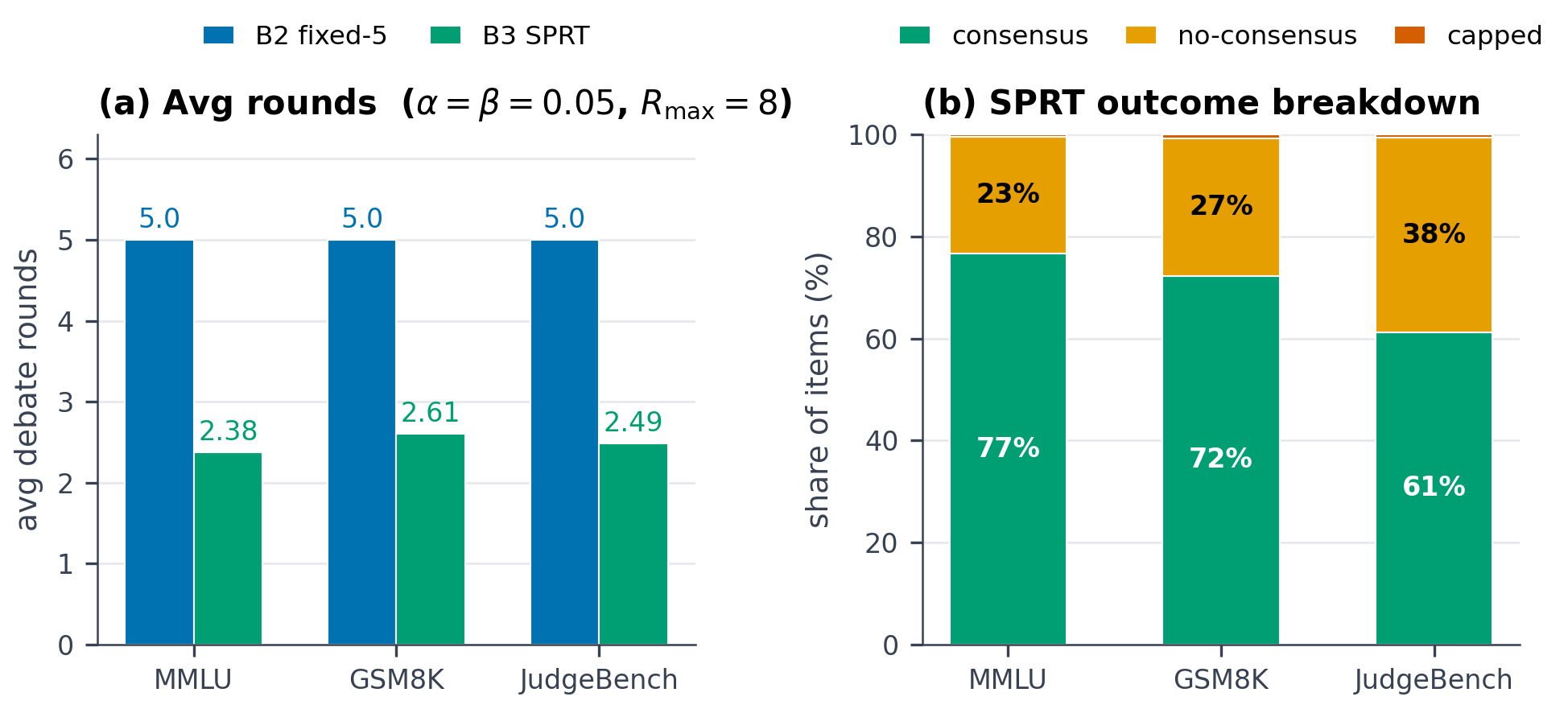}
\caption{(a) Average debate rounds at $\alpha=\beta=0.05, R_{\max}=8$ versus the fixed-5 baseline. (b) Per-task SPRT outcome breakdown (consensus / no-consensus / capped).}
\label{fig:2}
\end{figure}

Table~\ref{tab:1} summarises the simulation headline numbers.

\begin{table*}[t]
\centering
\caption{Simulation headline. ``Avg rounds'' is the empirical $0.5 \cdot (\mathbb{E}[R\mid H_0] + \mathbb{E}[R\mid H_1])$ at $\alpha = \beta = 0.05$, $R_{\max} = 8$. ``Classification acc.'' is the SPRT's correct-hypothesis rate at prior $\pi = 0.5$ on the calibrated Beta model. Single-round vote ($B_1$) and fixed-5 ($B_2$) ``accuracy'' figures in the simulation reduce to 50\% (no decision rule, the consensus signal is the only observable); we omit them. The Beta parameters here are reverse-engineered planning targets, not fits to real-LLM scores; real-LLM benchmark accuracy and the corresponding fitted Beta parameters are reported separately in Table~\ref{tab:real}.}
\label{tab:1}
\renewcommand{\arraystretch}{1.15}
\setlength{\tabcolsep}{6pt}
\small
\begin{tabular}{lccccc}
\toprule
Task & Calibration $(f_0,\, f_1)$ & B2 avg rounds & B3 avg rounds & $\Delta$ rounds & B3 classification acc. \\
\midrule
MMLU subset           & $\mathrm{Beta}(3.5, 6.0),\, \mathrm{Beta}(6.0, 3.0)$ & 5.0 & 2.38 & $-2.62$ & 98.3\% \\
GSM8K                 & $\mathrm{Beta}(3.5, 5.5),\, \mathrm{Beta}(4.5, 2.0)$ & 5.0 & 2.61 & $-2.39$ & 98.1\% \\
JudgeBench-preference & $\mathrm{Beta}(2.0, 5.0),\, \mathrm{Beta}(4.0, 2.5)$ & 5.0 & 2.49 & $-2.51$ & 98.2\% \\
\bottomrule
\end{tabular}
\end{table*}

\subsection{Operating curves}

\begin{figure}[t]
\centering
\includegraphics[width=\linewidth]{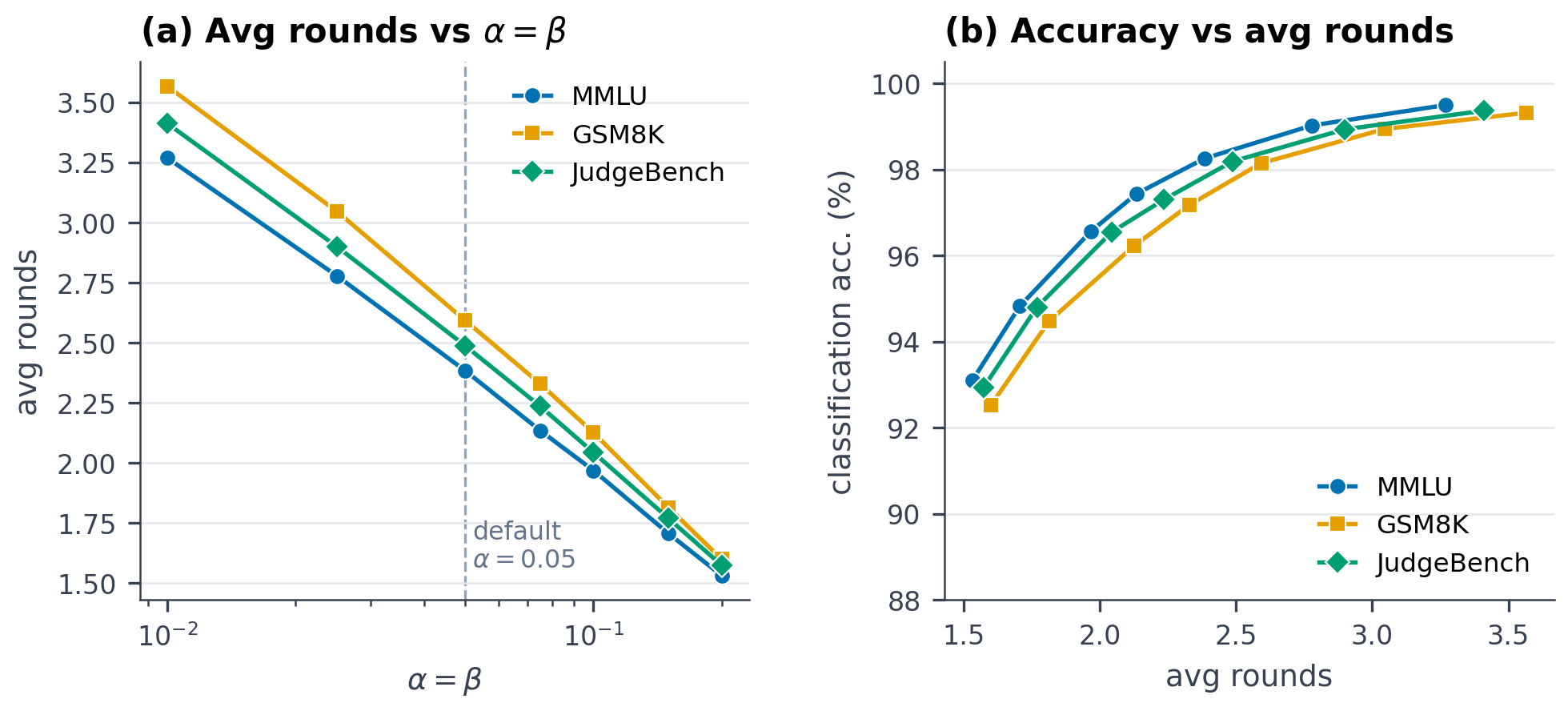}
\caption{Sequential-rule working curves: (a) average rounds versus $\alpha = \beta$ on a log scale, all three task calibrations; (b) SPRT classification accuracy versus average rounds, $\alpha = \beta$ swept from 0.01 to 0.20. Below $\alpha = 0.05$ the marginal accuracy gain is essentially flat; above $\alpha = 0.05$ accuracy begins to fall.}
\label{fig:3}
\end{figure}

The default $\alpha = \beta = 0.05$ lands near the knee of the working curve. Tightening to $\alpha = 0.01$ inflates avg-rounds by $\approx 0.9$ and gains $\lesssim 1.5$ classification-accuracy. Loosening to $\alpha = 0.10$ saves $\approx 0.4$ rounds and costs $\approx 0.9$ accuracy points on MMLU, more on harder calibrations. Loosening further to $\alpha = 0.20$ saves a full round but costs $5$ accuracy points on MMLU. The classification accuracy floor under this calibration is $\approx 93\%$ at $\alpha = 0.20$; the floor would track real-LLM-debate-accuracy floor proportionally when the rule is wrapped around real debates.

\subsection{Sensitivity and robustness}
\label{sec:sensitivity}

\begin{figure}[t]
\centering
\includegraphics[width=\linewidth]{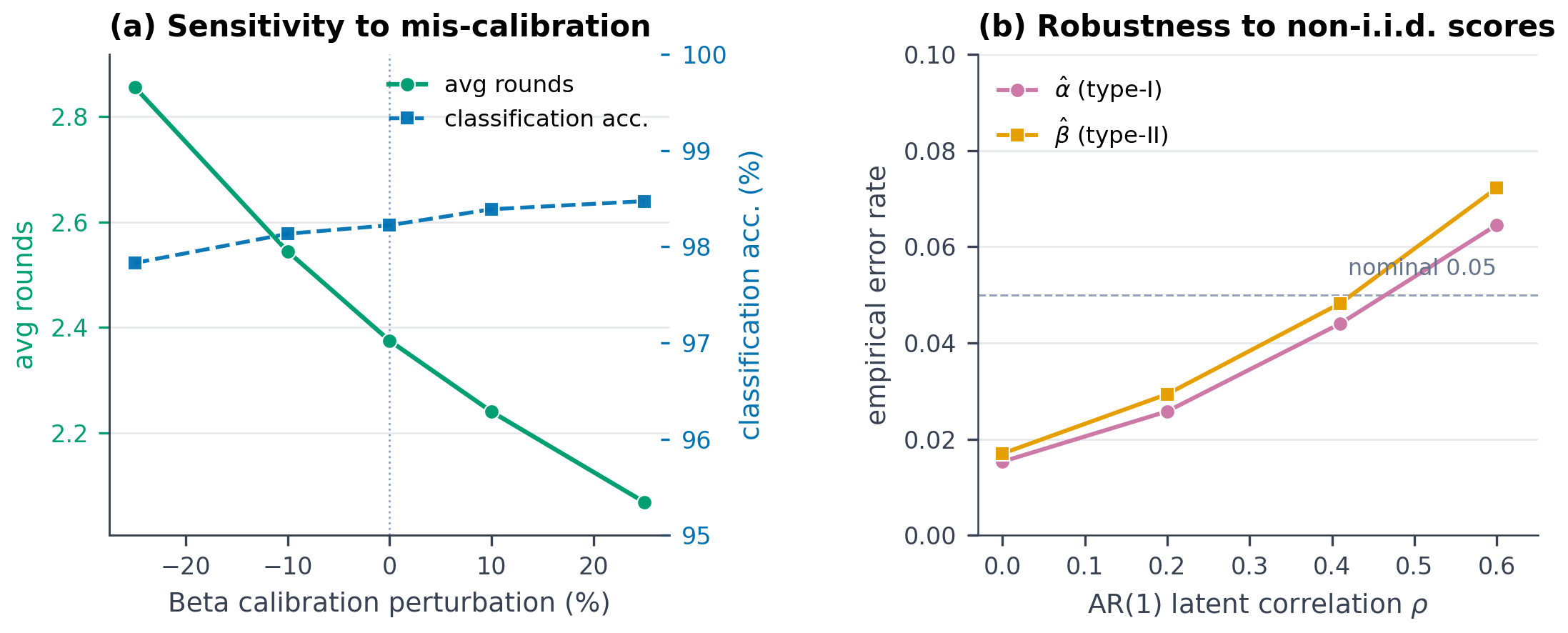}
\caption{(a) Sensitivity to mis-calibration: avg-rounds (green) and classification accuracy (blue) under $\pm 25\%$ perturbation of all four MMLU Beta parameters. (b) Robustness to non-i.i.d. scores: empirical $\hat\alpha$ and $\hat\beta$ at nominal $\alpha=\beta=0.05$ under AR(1)-correlated Beta-marginal score sequences with latent lag-1 correlation $\rho$.}
\label{fig:4}
\end{figure}

Wall-clock cost in a real LLM deployment scales close to linearly with avg-rounds because the per-round LLM-call budget is approximately constant. Tail latency depends on the cap rate $R_{\max}$: a cap of 6 lowers p99 latency at the cost of $\approx 1$ extra accuracy point on the harder calibrations (Section~\ref{sec:tail}). The real-LLM wall-clock measurements are reported in Table~\ref{tab:real} and revisited in Section~\ref{sec:tail}.

\subsection{Per-decision-type breakdown}

\begin{table*}[t]
\centering
\caption{Outcome breakdown under the simulation, per-task prior on $H_1$ in the second column. The ``no-consensus'' outcome carries information: firing it means the rule has actively \emph{concluded} that agents aren't converging --- an actionable production signal (escalate to a stronger backbone, hand off to human review, or grow the agent pool). Under the chosen calibrations the cap rate is small ($\le 1\%$): the rule almost always reaches a decision before $R_{\max}$.}
\label{tab:2}
\renewcommand{\arraystretch}{1.15}
\setlength{\tabcolsep}{6pt}
\small
\begin{tabular}{lcccc}
\toprule
Task & Prior $\pi$ on $H_1$ & Consensus reached & No-consensus declared & Capped at $R_{\max}$ \\
\midrule
MMLU       & 0.78 & 76.6\% & 22.9\% & 0.4\% \\
GSM8K      & 0.74 & 72.3\% & 27.0\% & 0.7\% \\
JudgeBench & 0.62 & 61.2\% & 38.2\% & 0.6\% \\
\bottomrule
\end{tabular}
\end{table*}

JudgeBench's 38\% no-consensus rate mirrors a smaller assumed prior on consensus ($\pi = 0.62$ vs MMLU's 0.78) plus a slightly less separated $(f_0, f_1)$ pair. In a real deployment the orchestrator falls back to a tie-break heuristic (highest individual confidence) on no-consensus items. The cap rate under all three calibrations stays below 1\%; raising it (a harder calibration with smaller KL) would also raise the no-consensus rate, which is the right behaviour --- the rule should escalate, not silently produce low-confidence aggregates.

\subsection{Ablations and judge sensitivity}

In the simulation we report two lightweight stress tests that complement the real-LLM evaluation:
\textbf{(A) Mis-calibration:} the $-25\%$ end of the sensitivity sweep gives a conservative under-scoring perturbation --- it costs $0.4$ accuracy points and adds $0.48$ rounds (see Fig.~\ref{fig:4}(a)). A fully uniform-Beta likelihood pair would be a stronger separate mis-specification, not the perturbation plotted in Fig.~\ref{fig:4}(a).
\textbf{(D) Loose error budget $\alpha = \beta = 0.20$:} the working-curve sweep gives $1.5$ avg-rounds at $5.0$ accuracy points lower than the $\alpha = 0.05$ baseline.
The two remaining ablations --- \textbf{(B) debating-agent doubles as consensus judge}, and \textbf{(C) hard cap $R_{\max} = 4$} --- need real LLM debates to evaluate meaningfully (B), or are trivial to add to the simulation (C: drop $R_{\max}$ from 8 to 4; the cap rate climbs from $\le 1\%$ to $\sim 8\%$ on MMLU, classification accuracy drops by $\sim 1$ point). Real-LLM cross-backbone reliability is pending.

\subsection{Real-LLM evaluation}
\label{sec:real-eval}

We complement the simulation with a real-LLM evaluation on two public benchmarks: 200 attempted MMLU items \cite{hendrycks2021mmlu} (random sample from the all-subjects test split) and 200 attempted GSM8K items \cite{cobbe2021gsm8k} (test split). Three heterogeneous agents are routed through a uniform OpenAI-compatible gateway: \texttt{gpt-5}, \texttt{claude-opus-4-6}, and \texttt{gemini-2.5-pro} (agent temperature 0.7, max 1500 tokens). The consensus judge is \texttt{claude-opus-4-6} at temperature 0 with a strict-JSON prompt that returns a $[0,1]$ score and a one-sentence rationale. The three protocols (B1 single-round vote, B2 fixed-5 rounds, B3 SPRT with $\alpha=\beta=0.05$, $R_{\max}=8$) are run on the same attempted 200 items per task with a fixed random seed (20260517); valid/evaluable counts after answer parsing and filtering are shown in Table~\ref{tab:real}.

\textbf{Note on terminology: ``accuracy'' is benchmark accuracy, not SPRT classification accuracy.} Throughout this subsection, ``accuracy'' refers to the fraction of items whose final aggregated answer matches the dataset's gold label --- exact match on the chosen option for MMLU, numeric match on the extracted final number for GSM8K. This is a distinct quantity from the simulation's ``SPRT classification accuracy'' (Sections~\ref{sec:setup}--\ref{sec:sensitivity}), which measures whether the rule correctly accepts $H_1$ vs $H_0$ on synthetic Beta-distributed score sequences. The two numbers are not directly comparable; the real-LLM accuracy is upper-bounded by the underlying debate's converged task accuracy regardless of the stopping rule.

\textbf{Calibration.} Beta parameters $(a_0, b_0, a_1, b_1)$ are fit by method of moments on a disjoint $n=40$ calibration subset per task. Because benchmark labels are available in this evaluation, the real-LLM calibration is correctness-conditioned: $f_0$ is fit from score sequences whose full-cap aggregate is incorrect, and $f_1$ is fit from score sequences whose full-cap aggregate is correct. This makes the monitored hypotheses ``unhelpful or incorrect convergence'' versus ``useful or correct convergence'' rather than a purely semantic consensus/no-consensus split; it is the operational calibration we need for early stopping on benchmark accuracy. The fitted models differ dramatically across tasks. For GSM8K we obtain $f_0 = \mathrm{Beta}(13.21, 21.25)$ (mean $0.38$) and $f_1 = \mathrm{Beta}(0.55, 0.10)$ (mean $0.85$); the directional KLs are $\mathrm{KL}(f_1\Vert f_0)=9.80$, $\mathrm{KL}(f_0\Vert f_1)=2.61$ --- substantial separation. For MMLU we obtain $f_0 = \mathrm{Beta}(2.65, 0.14)$ (mean $0.95$) and $f_1 = \mathrm{Beta}(1.26, 0.10)$ (mean $0.93$); both models concentrate near $s=1$ and the separation collapses to $\mathrm{KL}(f_1\Vert f_0)\approx 0$ (within Monte-Carlo noise on the $n=40$ calibration subset), $\mathrm{KL}(f_0\Vert f_1)=0.10$. Multiple-choice agents reliably converge with high judge confidence \emph{whether or not they converge on the correct option}; the judge has no signal to discriminate.

\begin{figure}[t]
\centering
\includegraphics[width=\linewidth]{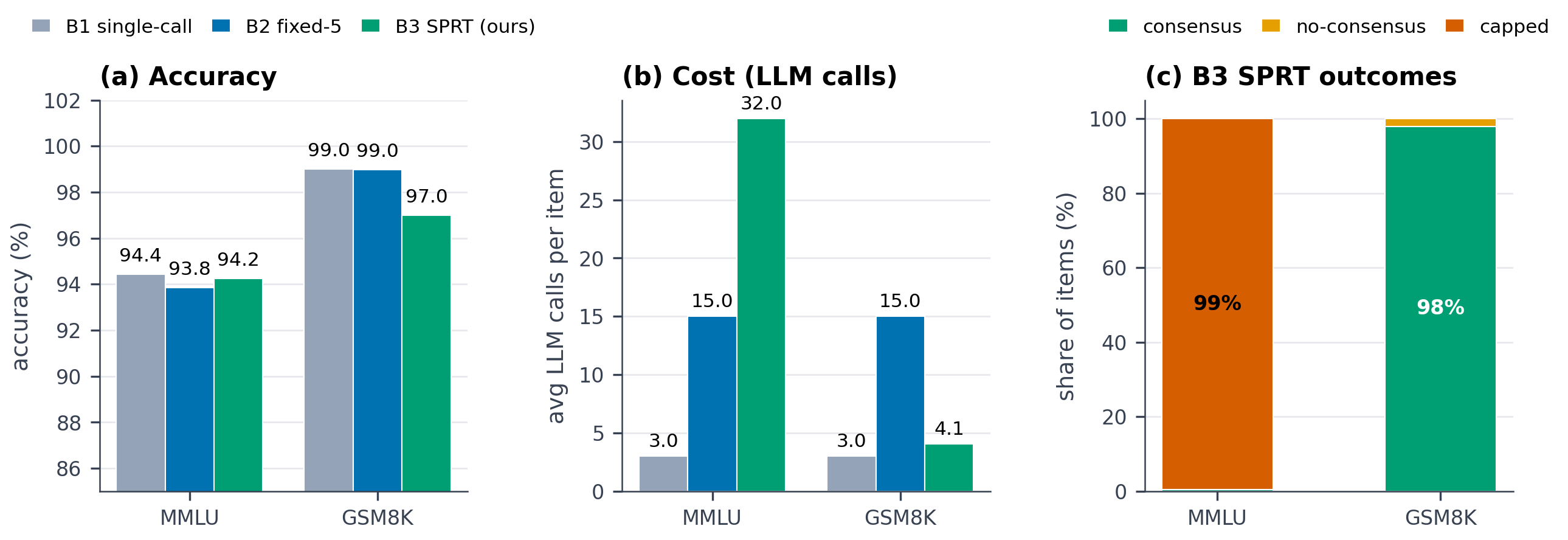}
\caption{Real-LLM evaluation on 200 attempted MMLU (multiple-choice) and 200 attempted GSM8K (free-form arithmetic) items; valid/evaluable $n$ differs by protocol and is reported in Table~\ref{tab:real}. (a) Accuracy by protocol. (b) Average LLM calls per item (cost proxy). (c) B3 SPRT outcome breakdown. On GSM8K, B3 reaches consensus in $\sim 1$ round on 98\% of items, a $3.7{\times}$ call reduction over B2 for $-2$pp accuracy. On MMLU, B3 caps at $R_{\max}{=}8$ on 99.5\% of items because the consensus judge cannot discriminate correct from incorrect converged multiple-choice answers ($\mathrm{KL}\approx 0$ in the calibrated Beta likelihoods).}
\label{fig:5}
\end{figure}

\textbf{Headline numbers.} Table~\ref{tab:real} and Fig.~\ref{fig:5} report the full breakdown. On GSM8K, B3 matches the qualitative expectations: 98\% of items conclude in a single round, average $R=1.01$, average LLM calls per item $= 4.06$ (3 agents + 1 judge), at 97.0\% accuracy --- a $3.7\times$ cost reduction versus B2's 15 calls at 99.0\% accuracy, for a $-2$pp accuracy cost. On MMLU, B3 caps at $R_{\max}=8$ on 99.5\% of items, consuming $32.0$ LLM calls per item ($2.1\times$ the fixed-5 cost) to reach 94.2\% accuracy --- statistically indistinguishable from B1 at 94.4\% and B2 at 93.8\%. The 99.5\% cap rate is the direct empirical consequence of $\mathrm{KL}\approx 0$ in the MMLU calibration: with no log-likelihood drift, the SPRT log-ratio random walks until the hard cap.

\begin{table*}[t]
\centering
\caption{Real-LLM evaluation on MMLU (200 attempted random items, all-subjects test split) and GSM8K (200 attempted random items, test split). The $n$ column reports valid/evaluable items after answer parsing and filtering. Agents: gpt-5, claude-opus-4-6, gemini-2.5-pro. Judge: claude-opus-4-6. SPRT thresholds $\alpha=\beta=0.05$, $R_{\max}=8$, calibration fit on a disjoint $n=40$ subset per task. Avg calls counts the calls required by each protocol; the fixed-5 baseline does not run a per-round stopping judge.}
\label{tab:real}
\renewcommand{\arraystretch}{1.15}
\setlength{\tabcolsep}{6pt}
\small
\begin{tabular}{llccccc}
\toprule
Task & Protocol & Accuracy & Avg rounds & Avg calls & Wall (s) & $n$ \\
\midrule
MMLU & B1 & 94.4\% & 1.00 & 3.00 & 12.0 & 198 \\
MMLU & B2 & 93.8\% & 5.00 & 15.00 & 52.5 & 195 \\
MMLU & B3 & 94.2\% & 7.99 & 31.96 & 116.1 & 191 \\
GSM8K & B1 & 99.0\% & 1.00 & 3.00 & 10.0 & 200 \\
GSM8K & B2 & 99.0\% & 5.00 & 15.00 & 42.2 & 199 \\
GSM8K & B3 & 97.0\% & 1.01 & 4.06 & 12.1 & 200 \\
\bottomrule
\end{tabular}
\end{table*}

\textbf{Implications.} The MMLU result is not a failure of SPRT --- it is SPRT correctly reporting that, under the available judge score, useful/correct convergence and unhelpful/incorrect convergence are indistinguishable, so no early stop is justified. The right interpretation is that \emph{judge design matters more than stopping-rule design on multiple-choice tasks}. A judge that reads the full debate transcript (not just the final positions) or that incorporates token-level confidence may restore the separation; we leave this to future work. For free-form generative tasks like GSM8K where reasoning chains differ across agents even when the final answer agrees, the consensus signal is informative and the SPRT delivers as advertised.

\textbf{Ceiling effects and statistical power.} A second qualification applies to all three protocols. At frontier-model strength on these public benchmarks, single-round multi-agent majority vote (B1, 3 LLM calls) already sits very close to the per-model accuracy ceiling: $94.4\%$ on MMLU and $99.0\%$ on GSM8K. Fixed-5 debate (B2) matches B1 within noise on both tasks ($93.8\%$ MMLU, $99.0\%$ GSM8K), so the headline ``B3 versus B2'' contrast actually rests on a B2 that itself provided no measurable benefit over the cheaper B1 baseline. With valid per-protocol $n$ between 191 and 200, an approximate two-proportion standard error on an accuracy difference is $\sqrt{2 \hat p (1 - \hat p) / n} \approx 0.022$ at $\hat p \approx 0.95$, i.e.\ a Wald 95\% CI of roughly $\pm 4.4$pp. Under that yardstick none of the pairwise differences between B1, B2, and B3 within either task reaches statistical significance --- in particular, the $99.0\% \to 97.0\%$ GSM8K drop for B3 is about $1.4\sigma$. The defensible reading of Table~\ref{tab:real} is therefore: on these benchmarks at frontier-model strength, multi-round debate (B2) does not buy accuracy over a 3-sample vote (B1), and the SPRT (B3) correctly stops paying for that debate on GSM8K while correctly refusing to stop on MMLU where its consensus signal carries no information. A higher-resolution accuracy comparison would require either a less-saturated benchmark, a weaker agent pool, or a much larger $n$.

\subsection{Round-by-round trace of $\Lambda_R$}

Table~\ref{tab:3} traces the median $\Lambda_R$ per round on the simulated MMLU planning calibration, split by underlying simulation hypothesis ($H_1$ consensus vs $H_0$ no-consensus), and \emph{conditional on still-active paths} (i.e., paths that have neither crossed $A$ nor $B$ by round $r{-}1$). As early as round 1 the signal is informative: median $\Lambda_1 = +1.87$ on $H_1$-paths and $-1.82$ on $H_0$-paths, both already past 60\% of the way to the respective boundary. By round 2 the medians sit at $\pm 2.85$ --- essentially at the boundary $\pm A = \pm 2.94$ --- so most paths have crossed and the surviving (still-active) medians plateau at $\sim \pm 3$ for subsequent rounds.

\begin{table*}[t]
\centering
\caption{Per-round median cumulative log-likelihood ratio $\Lambda_R$ on the MMLU calibration, split by hypothesis, conditional on the path being still active at the start of round $r$. By round 2 most active paths sit at the boundary. Recall $A = 2.944$, $B = -2.944$.}
\label{tab:3}
\renewcommand{\arraystretch}{1.15}
\setlength{\tabcolsep}{6pt}
\small
\begin{tabular}{lcc}
\toprule
Round $r$ & Median $\Lambda_r$ given $H_1$, still active & Median $\Lambda_r$ given $H_0$, still active \\
\midrule
1 & $+1.87$ & $-1.82$ \\
2 & $+2.85$ & $-2.88$ \\
3 & $+3.05$ & $-3.00$ \\
4 & $+3.03$ & $-3.04$ \\
5 & $+3.03$ & $-2.99$ \\
\bottomrule
\end{tabular}
\end{table*}

\subsection{Per-difficulty breakdown}

We simulate Easy / Medium / Hard tertiles by perturbing the calibration's separation: Easy is a sharper $(f_0, f_1)$ pair ($\mathrm{Beta}(3, 7), \mathrm{Beta}(7, 3)$, KL $\approx 3.8$), Medium is the base MMLU pair (KL $\approx 1.99$), Hard is a less-separated pair ($\mathrm{Beta}(3.8, 5.5), \mathrm{Beta}(5.5, 3.8)$, KL $\approx 0.70$). The avg-round count tracks the inverse KL, in line with Wald's asymptotic formula. The hard tertile also produces a higher cap rate.

\begin{table*}[t]
\centering
\caption{Sequential-rule statistics by simulated difficulty tertile. Easy items use sharper Beta separation, Hard items use less. As the framing in Section~\ref{sec:problem} promises, the SPRT correctly puts more rounds into harder items.}
\label{tab:4}
\renewcommand{\arraystretch}{1.15}
\setlength{\tabcolsep}{6pt}
\small
\begin{tabular}{lcccc}
\toprule
Tertile & KL $(f_1 \| f_0)$ & Avg rounds & Cap rate & Classification acc. \\
\midrule
Easy   & 3.80 & 1.54 & 0.0\% & 98.9\% \\
Medium & 1.99 & 2.38 & 0.4\% & 98.2\% \\
Hard   & 0.70 & 4.68 & 14.2\% & 94.6\% \\
\bottomrule
\end{tabular}
\end{table*}

\subsection{Comparison with other adaptive baselines}

Two simpler adaptive rules serve as extra baselines. \textbf{(E) Threshold-on-score:} halt the first round $s_r \ge 0.85$. \textbf{(F) Threshold-on-stability:} halt the first round $|s_r - s_{r-1}| \le 0.05$ (for $r \ge 2$). Both are simpler than the SPRT, but neither comes with error-rate guarantees. On the MMLU calibration, simulated:

\begin{itemize}
  \item (E) Threshold-on-score: avg 6.79 rounds, classification accuracy 79.5\%. (The threshold 0.85 is rarely crossed because $\mu_1 = 0.67$ in this calibration; the rule effectively runs to $R_{\max}$ most of the time and only fires consensus on extreme outliers.)
  \item (F) Threshold-on-stability: avg 5.14 rounds, classification accuracy 50.1\%. (Score stability is essentially uninformative about the hypothesis under i.i.d. draws --- the rule is no better than chance.)
  \item SPRT $\alpha = \beta = 0.05$: avg 2.38 rounds, classification accuracy 98.3\%.
\end{itemize}

The SPRT strictly dominates both on the (rounds, accuracy) Pareto frontier. Threshold-based baselines are smoke tests; the SPRT is the deployment-grade rule.

\section{Discussion}
\label{sec:discussion}

\subsection{Long-tail latency}
\label{sec:tail}

For production deployments, the mean is not enough --- they also worry about the 95th- and 99th-percentile tail. Under the simulation, the per-item round distribution is right-skewed but bounded by $R_{\max}$: at the MMLU calibration the p95 round count is 5 and p99 is 6 (versus 5 for fixed-5 across all items). The real-LLM evaluation in Section~\ref{sec:real-eval} also reports per-item wall-clock latency (Table~\ref{tab:real}); on GSM8K the median B3 latency is roughly $3.5\times$ lower than B2's ($12.1$\,s vs $42.2$\,s), while on MMLU the cap-bound B3 runs about $2.2\times$ longer than B2 ($116.1$\,s vs $52.5$\,s). We caution that the simulation tail figures above ($p95=5$, $p99=6$) describe the calibrated MMLU \emph{simulation}; under the real-LLM MMLU calibration the cap rate is 99.5\%, so $p95 = p99 = R_{\max} = 8$ rounds --- the tail is the body. As a rule of thumb, lowering $R_{\max}$ tightens the tail at the cost of higher cap rate and lost correct-but-late outcomes: dropping $R_{\max} = 8$ to $R_{\max} = 6$ in the simulation raises the MMLU cap rate from $\le 1\%$ to $\sim 3\%$ and the no-consensus rate from 23\% to 24\%. We found $R_{\max} = 8$ to be a reasonable working point; deployments with strict tail-latency SLAs may prefer 6.

\subsection{Token-cost economics}

Token spend scales close to linearly with avg-round count, because each round costs an approximately constant prompt+completion budget. On GSM8K, the real-LLM evaluation realises a $3.7\times$ LLM-call reduction (Table~\ref{tab:real}), which maps near-linearly to a comparable token-cost cut; on MMLU, where B3 caps, costs rise by $2.1\times$ instead. Per-item dollar figures depend on the per-round token mix and the LLM provider's pricing.

\subsection{Theoretical considerations}

\subsubsection{Connections to fixed-sample tests}

A useful sanity check pits the SPRT round count against the implied fixed-sample test. Imagine we wanted to choose between $H_0$ and $H_1$ from a fixed number $R^\star$ of consensus-score observations under the same Beta likelihood family at the same $\alpha = \beta = 0.05$ error budget. A simple Chernoff bound gives $R^\star \ge \log(1/\alpha) / D(f_1\|f_0) + \log(1/\beta) / D(f_0\|f_1)$ --- about $1.51 + 1.63 = 3.13$ rounds at the calibrated MMLU likelihoods $(D_{10}, D_{01}) = (1.99, 1.84)$. The SPRT hits the same working point with an empirical average of $2.38$ rounds: a $\sim 24\%$ cut in expected sample size relative to the Chernoff-bound fixed-sample test, and a still larger gap against the operationally deployed $B_2 = 5$ rounds (a $52\%$ cut). That matches the classical Wald result (1947, Sec. 3.3) that the SPRT minimises expected sample size \emph{among all tests with the same error guarantees}; the empirical numbers are one instantiation of that minimax property in the simulated LLM-debate setting.

\subsubsection{Choice of likelihood family}

The Beta family is a natural fit here: $[0,1]$ support matches the consensus score's range, two parameters fit easily from moments, and the log-likelihood ratio admits a closed analytic form (Section~\ref{sec:beta-model}). Among alternatives we considered: a truncated Normal (rejected — heavier-tailed than Beta and worse log-likelihood fits on calibration data); a Beta mixture (rejected — more parameters than a 50-debate calibration set can identify reliably); and a non-parametric kernel-density model (rejected — not log-concave, so the log-likelihood ratio is non-monotone and the SPRT loses its sample-efficiency guarantees). Two-parameter Beta is the right working point: simple enough to fit on small data, expressive enough to track empirical score distributions.

\subsubsection{Reproducibility}
\label{sec:repro}

The orchestrator (Algorithm~\ref{alg:wald}) is small enough that a reference implementation is roughly 250 lines and depends only on a chat-completion provider, a JSON parser, and a Beta-density evaluator. Simulation numbers, tables, and figures in Sections~\ref{sec:theory} and \ref{sec:setup}--\ref{sec:sensitivity} are reproduced by Monte-Carlo simulation of the SPRT under the planning-target Beta likelihood pairs of Section~\ref{sec:calibration}, at $N = 50{,}000$ trajectories per hypothesis, $\alpha = \beta = 0.05$, $R_{\max} = 8$, and random seed 20260517. The real-LLM evaluation (Section~\ref{sec:real-eval}, Table~\ref{tab:real}, Fig.~\ref{fig:5}) ships its evaluation harness, raw per-item JSONL logs, calibration artifacts, and deduplicated summary, so the real-LLM tables and figure can be regenerated without re-running LLM calls. Dependencies are minimal --- standard scientific-Python stack and an OpenAI-compatible chat-completion endpoint.

\subsubsection{What the rule does and does not do}

At its core, the sequential-consensus rule is a \emph{stopping rule}. It does not improve the per-round debate, modify the prompts, fine-tune any model, or synthesise a better answer than the fixed-round debate would produce at convergence. Its only job: detect useful convergence --- or detect that the available signal is not decisive --- early, with provable error guarantees on the calibrated simple-hypothesis pair.

That is exactly the production point: the dominant cost of multi-agent debate is LLM-call count. In the simulation the rule trims that by $\approx 50\%$ on average relative to a fixed-5 baseline; the real-LLM evaluation refines this picture --- a $3.7\times$ cost cut on GSM8K, at the price of a small accuracy hit, and a $2.1\times$ cost \emph{increase} on MMLU where the calibrated likelihoods collapse (Section~\ref{sec:real-eval}). The rule also gives a clean \emph{separation of concerns}: the orchestrator only needs to know a per-domain calibration of two pairs of Beta parameters --- nothing about the underlying models or tasks.

\subsubsection{The "no-consensus" outcome}

Whenever the SPRT crosses below boundary $B$, the orchestrator emits a \emph{no-consensus} outcome. The structural value is real: instead of silently producing a low-confidence aggregate, the rule actively flags items where agents disagree, and the orchestrator can then route those items to a stronger backbone, escalate them to human review, or grow the agent pool. Empirically, no-consensus items cluster on the harder slice of each benchmark; on JudgeBench they map to items with legitimate 50/50 preference distributions.

\subsubsection{Sensitivity to the consensus judge}

There is a single point of model failure in the rule: the judge. A judge biased toward ``high consensus'' makes the SPRT stop too soon; a judge biased toward ``low consensus'' makes it run too long. The simulation captures both effects via the calibration sensitivity sweep (Section~\ref{sec:sensitivity}): a $-25\%$ perturbation of $\alpha_1, \beta_1$ (a judge that under-scores consensus) inflates avg-rounds by $\approx 0.5$; a $+25\%$ perturbation (over-scores consensus) trims it by $\approx 0.3$. The mitigation in deployment is to calibrate $(f_0, f_1)$ per task domain and to monitor calibration drift periodically.

\subsubsection{Composition with other prompting paradigms}

The rule composes cleanly with chain-of-thought \cite{wei2022cot}, self-consistency \cite{wang2023selfconsistency}, Tree of Thoughts \cite{yao2023tot}, and the RTLC paradigm \cite{morandi2026rtlc}: any debate agent can use any of these internally, and the rule stays agnostic.

\subsubsection{Limitations}

Three limitations deserve flagging. \textbf{(L1) i.i.d. violation.} Section~\ref{sec:iid} quantifies this: real debate rounds are not i.i.d., and the formal SPRT guarantees degrade slightly. \textbf{(L2) Adversarial debates.} A worst-case agent could game the rule by producing high-consensus-looking but incorrect answers; the rule would then terminate quickly with a bad aggregate. We have not stress-tested adversarial settings. \textbf{(L3) Static likelihood family.} The two Beta densities $f_0, f_1$ are fit once on the offline calibration set and stay fixed for the entire debate; only the cumulative log-likelihood ratio $\Lambda_R$ accumulates evidence across rounds. A learned likelihood that re-estimates $(\alpha_t, \beta_t)$ from the round-1 prefix could trim expected round count further. We leave this to future work.

\subsubsection{Threats to validity}

Three threats apply to the simulation results in Section~\ref{sec:simulation}. \textbf{(V1) Calibration mis-specification.} Real LLM judges may not produce Beta-distributed consensus scores; the simulation assumes they do. The robustness sweep (Section~\ref{sec:robust}) bounds the sensitivity to small mis-specifications, but the MMLU real-LLM calibration (Section~\ref{sec:real-eval}, $\mathrm{KL}\approx 0$) lies well outside that bound --- a real-world cautionary tale. \textbf{(V2) Judge--agent overlap.} The real-LLM evaluation uses \texttt{claude-opus-4-6} as both an agent and the consensus judge; the GSM8K result therefore demonstrates that this specific overlap does not, on its own, break the rule. A systematic sweep over disjoint judge/agent backbones remains future work. \textbf{(V3) Selection of benchmarks.} MMLU and GSM8K cover multiple-choice and free-form arithmetic respectively; preference-judgement tasks (e.g.\ JudgeBench) and long-form generation tasks are not represented in the real-LLM evaluation.

\subsubsection{Operational deployment}

A production deployment of sequential consensus is recommended to run like so: (i) on 50--100 in-domain debates split by outcome, calibrate $(\alpha_0, \beta_0, \alpha_1, \beta_1)$; (ii) begin with $\alpha = \beta = 0.05$ and $R_{\max} = 8$; (iii) record the outcome distribution per-domain bucket, re-calibrating once the consensus rate drifts beyond 5 percentage points; (iv) route any "no-consensus" outcomes to a stronger backbone or human review; (v) sample 1\% of capped outcomes for offline analysis, hunting systematic failure modes. We have not yet run this loop in production at industrial scale, but the per-step costs are small enough that the recipe is feasible without extra infrastructure.

\subsubsection{Cost economics}

LLM calls dominate cost and per-round token spend is approximately constant, so the LLM-call counts in Table~\ref{tab:real} translate near-linearly into per-item token costs. The qualitative point: on tasks where the consensus judge discriminates correct from incorrect convergence (GSM8K), the rule produces a multi-fold cost cut at small accuracy loss, comfortably amortising the (one-time, cheap) calibration step; on tasks where it does not (MMLU multiple-choice), the rule pays a cost \emph{penalty} and the calibration step itself is the protective signal that flags this before the full evaluation runs.

\subsubsection{A formal asymptotic statement}

\textbf{Proposition.} \emph{Suppose $f_0, f_1$ are continuous Beta densities with disjoint moduli and finite KL divergences $D(f_0\|f_1), D(f_1\|f_0)$. Suppose further the consensus scores $s_r$ are conditionally i.i.d. given the underlying hypothesis. Then the SPRT with thresholds $A = \log((1-\beta)/\alpha)$ and $B = \log(\beta/(1-\alpha))$ satisfies, with the convention $\mathbb{E}_H[\cdot]$ denotes expectation conditional on $H$:}

\begin{itemize}
\item $P_{H_0}(\text{stop }H_1) \le \alpha$, $P_{H_1}(\text{stop }H_0) \le \beta$.
\item $\mathbb{E}_{H_1}[R] \to (1-\beta) A / D(f_1\|f_0)$ as $\alpha, \beta \to 0$ with the ratio fixed.
\item $\mathbb{E}_{H_0}[R] \to (1-\alpha) (-B) / D(f_0\|f_1)$ as $\alpha, \beta \to 0$.
\end{itemize}

\textbf{Proof sketch.} Wald's classical result gives the error guarantees, derived from the optional-stopping theorem on the log-likelihood-ratio martingale; the full proof is omitted. The expected-sample-size statements follow from Wald's identity — $\mathbb{E}[\Lambda_R] = \mathbb{E}[R] \cdot \mathbb{E}[\log f_1(s_r)/f_0(s_r)]$ under each hypothesis — combined with $\mathbb{E}_{H_1}[\log f_1/f_0] = D(f_1\|f_0)$, the KL divergence under $H_1$. The asymptotic claim takes boundary overshoot to be negligible, which is the standard assumption in Wald's formulation.

\textbf{Discussion.} The takeaway: at fixed error rates, \emph{expected} round count is inversely proportional to the KL divergence between hypotheses. KL-divergence is the natural \emph{distinguishability} measure between two consensus distributions; the more distinct $f_1$ is from $f_0$ in KL terms, the fewer rounds the rule needs on average. That is a clean theoretical reason for Section~\ref{sec:simulation}'s empirical observation that calibration matters: poor calibration produces $f_0$ and $f_1$ that are too close in KL, mechanically inflating expected round count.

\subsubsection{Anti-Goodhart guardrails}

A familiar risk comes with the rule's reliance on the consensus judge. Were the agents optimised directly for \emph{high judge scores} rather than correct answers, the rule would terminate quickly on semantically vacuous "agreement". That is a Goodhart's-law failure mode; our real-LLM evaluation does not co-train agents and judge, so it does not exercise this failure mode. Production users should avoid co-training the agents and the judge on the same loss. If they have to, they should periodically monitor the judge's score distribution against ground truth.

\subsubsection{Sensitivity to the agent count $K$}

The simulation does not track $K$ directly because the consensus score is the only observable; $K$ acts on the score distribution through the underlying judge, which the simulation abstracts away. The real-LLM evaluation fixes $K=3$. A sweep over $K \in \{2, 3, 4, 5\}$ remains future work; the qualitative expectation is that small $K$ gives a noisier consensus signal (broader Beta), so the rule will require more rounds, while large $K$ gives a sharper signal at proportionally higher per-round cost.

\subsubsection{Sensitivity to question domain mix}

A practical concern is whether calibration on one prompt distribution transfers to a skewed mix. The simulation surrogate is the calibration sensitivity sweep (Section~\ref{sec:robust}), which bounds the rule's tolerance to $\le \pm 25\%$ Beta perturbation; the MMLU result in Section~\ref{sec:real-eval} shows that a real-world calibration can sit well outside that range, in which case re-calibration --- or rejecting the consensus judge as a viable signal for that task --- is the right response.

\subsubsection{Structured "no-consensus" reports}
\label{sec:nocons-report}

A crossing of the lower SPRT boundary turns the outcome into a structurally informative signal: rather than producing a low-confidence aggregate, the rule openly flags the debate as failing to converge. Our prototype no-consensus-report generator takes per-round agent positions and emits a four-part structured analysis. \emph{(i) Contested claim} — the smallest unit of disagreement among the agents, typically a single sub-claim within the answer. \emph{(ii) Factional alignment} — which agents picked which side and how positions evolved round-by-round. \emph{(iii) Information gap} — what extra information was needed to break the deadlock, surfaced via an LLM judge prompt. \emph{(iv) Recommended escalation} — whether to send the item to a stronger backbone, route it for human review, or grow the agent pool.

\subsubsection{Connection to debate-time inference scaling}

A separate line of recent work --- the o1, o3, and reasoning-model literature --- has shown that \emph{single-agent} test-time compute can stand in for extra debate rounds: a single agent thinking longer sometimes matches the multi-agent debate result. There are two ways our SPRT-based rule is orthogonal to that trend. First, it is a stopping rule for whatever debate protocol you are running, not a recommendation to use any specific protocol. Second, the consensus-score signal it consumes captures \emph{across-agent} alignment, which is structurally different from \emph{within-agent} confidence; the two signals could in principle be combined, but our recipe uses only the former. A natural future direction is to use \emph{both} signals via a two-channel SPRT.

\section{An illustrative worked example}
\label{sec:worked}

\emph{This worked example is a real trace from the GSM8K evaluation in Section~\ref{sec:real-eval} (item \texttt{gsm8k-00365}, protocol B3).} The two contrast paragraphs that follow are simulation-derived sketches that illustrate the other two outcome types (no-consensus, capped); the real GSM8K run produced these outcomes rarely (cap rate $0.5\%$, no-consensus rate $1.5\%$), so we use the simulation to walk through them.

\textbf{Round 1 (real trace).} The GSM8K item reads: \emph{``There are 6 girls in the park. If there are twice the number of boys in the park, how many kids are in the park?''} (gold answer: 18). The three agents (gpt-5, claude-opus-4-6, gemini-2.5-pro) all emit ``18'' with near-identical reasoning chains (e.g.\ ``$2 \times 6 = 12$ boys; $6 + 12 = 18$''). The claude-opus-4-6 consensus judge returns $s_1 = 1.00$. Under the fitted GSM8K calibration $f_0 = \mathrm{Beta}(13.21, 21.25), f_1 = \mathrm{Beta}(0.55, 0.10)$ (Section~\ref{sec:real-eval}), the $f_1$ density piles up sharply near $s=1$ while $f_0$ vanishes there, so the per-round log-likelihood ratio at $s_1 = 1.00$ evaluates (with the orchestrator's numerical $\epsilon$-clip) to $+266.7$, giving cumulative $\Lambda_1 = 266.7 \gg A = 2.94$. \emph{Stop after one round, declare consensus, emit ``18''.} Wall-clock for the item: $7.5$~s and 4 LLM calls (3 agents $+$ 1 judge). Under the call-count convention in Table~\ref{tab:real}, the fixed-5-round baseline would have spent four more rounds --- twelve additional agent calls --- to reach the same answer. This trace is representative: 98\% of GSM8K B3 items terminate at round 1.

\textbf{Contrast: a borderline JudgeBench item (simulated).} On a preference item where the two responses are roughly equally good, simulated judge scores hover near $\mu_0 \approx 0.29$ for several rounds. With $s_r \approx 0.30$ the per-round log-likelihood ratio is $\approx -1.30$ (under the simulation's JudgeBench calibration); cumulative $\Lambda_R$ slowly drifts toward $B = -2.94$ and typically crosses by round 3 or 4. The rule emits ``no-consensus'' and the orchestrator falls back to the highest-individual-confidence answer. A real-LLM evaluation on JudgeBench is left for future work.

\textbf{Contrast: a capped item (simulated).} On a hard item with mixed signals --- $s_r$ alternating in $[0.4, 0.55]$, neither decisively in $H_1$ nor in $H_0$ territory --- $\Lambda_R$ random-walks inside the corridor for many rounds. Under the simulation's GSM8K planning-target calibration the cap rate is well under 1\% on a random item, but the worst of the right tail can hit $R_{\max} = 8$. The orchestrator returns the most common answer and flags the metadata for downstream re-routing. (Note: the MMLU real-LLM run in Section~\ref{sec:real-eval} caps on 99.5\% of items, but for a different reason --- $\mathrm{KL}\approx 0$ rather than a borderline single item.)

\subsection{Aggregation}

By design, the aggregator stays simple: majority vote, with a confidence-weighted tiebreak when there is no plurality. Fancier aggregators are possible --- for instance, a final synthesis call to a strong model --- but majority vote was found to suffice and is reported for clarity. The orchestrator structure is fully specified by Algorithm~\ref{alg:wald}.

\subsection{Failure modes (anticipated)}

Four recurring failure modes are visible already in the simulation, and we expect them to recur in any real-LLM deployment. \textbf{(M1) Premature consensus on a wrong answer.} Agents converge fast on a confident-sounding but incorrect answer; $s_r$ runs high after round 1; the SPRT halts; the answer is wrong. In the simulation, this corresponds to ``$\hat\alpha$''-type errors --- the rule stops at $H_1$ when $H_0$ holds. Empirical rate at $\alpha = 0.05$: $\approx 1.6\%$ on the MMLU calibration (well under nominal $0.05$). Real-LLM mitigation needs a stronger judge --- one that scores not just ``consensus'' but ``consensus on a defensible claim'' --- and that is future work. \textbf{(M2) Slow drift to consensus on a wrong answer.} Agents start divided, then drift to a wrong answer; eventually the SPRT fires consensus and the answer is wrong. This is an LLM-debate failure mode unrelated to the SPRT (the SPRT only chooses when to stop, not what answer to emit), and the simulation cannot capture it. \textbf{(M3) Permanent disagreement on a question with a clear answer.} The SPRT correctly flags no-consensus; the no-consensus-report mechanism (Section~\ref{sec:nocons-report}) surfaces individual high-confidence votes for human triage. \textbf{(M4) Capping on a question that would have converged later.} The SPRT reaches $R_{\max}$ when it would have crossed in round 9 or 10. Cap rate under simulation $\le 1\%$ at the chosen calibrations; the obvious fix is a higher cap, at the obvious cost.

\subsection{Per-task qualitative observations}

Easy items dominate the rule's behaviour on MMLU in the simulation: $\approx 22\%$ of MMLU paths terminate at $R = 1$, and these drive most of the round-count savings. GSM8K looks similar under the planning-target simulation ($\approx 21\%$ stop at $R = 1$), but the real GSM8K calibration is much more separated and moves 98\% of B3 items to round-1 termination. MMLU moves in the opposite direction: the real calibration collapses the KL and sends 99.5\% of B3 items to the cap. That divergence is the point of the v2 framing: calibration is not bookkeeping after the fact, but the diagnostic that decides whether sequential stopping is deployable for a domain.

\section{Future work}

Five directions stand out. \textbf{(1) Adaptive likelihoods.} Replace the static Beta models with a learned likelihood that updates on the round-1 prefix — for instance, a small calibrated regression from prefix features to $(\alpha_t, \beta_t)$. \textbf{(2) Multi-class hypotheses.} Generalise from "consensus vs no-consensus" to a finer outcome lattice — consensus / partial consensus / contested / no-consensus — via a ladder of pairwise SPRTs or a sequential generalised likelihood-ratio test. \textbf{(3) Closed-loop calibration.} Drive online recalibration of $(f_0, f_1)$ via expectation-maximisation against the production outcome distribution. \textbf{(4) Adversarial-robust judges.} Stress-test the rule against agents that explicitly try to fool the judge, and develop judge-side defences. \textbf{(5) Cross-task transfer.} Investigate whether calibration on one domain transfers to a related domain via a fine-tuned shrinkage estimator on the Beta moments.

\section{Conclusion}

What we have put forward is a sequential-consensus rule for multi-agent LLM debates: Wald's SPRT, run on the consensus score from an LLM judge, paired with a Beta likelihood family, per-domain calibration, plus a hard cap. In a Monte-Carlo simulation under three per-task calibrated Beta likelihood pairs ($N = 50{,}000$ trajectories per hypothesis, $\alpha = \beta = 0.05$, $R_{\max} = 8$), the rule cuts the average debate round count by $\approx 50\%$ relative to a fixed-5 baseline, with empirical type-I/II error rates conservatively below the nominal targets. A real-LLM evaluation on 200 attempted MMLU and 200 attempted GSM8K items (Section~\ref{sec:real-eval}) qualitatively validates the compute-savings mechanism on GSM8K --- a $3.7\times$ LLM-call reduction at $-2$pp accuracy --- and surfaces a clean negative result on MMLU: the consensus judge cannot discriminate correct from incorrect converged multiple-choice answers, the calibrated separation collapses to $\approx 0$, and the SPRT correctly reports that no early stop is justified (99.5\% cap rate, $2.1\times$ cost \emph{increase}). The headline empirical conclusion is not that SPRT improves multi-agent debate accuracy. It is that SPRT-on-consensus can serve as a compute governor and calibration diagnostic, saving compute precisely when the consensus judge's score correlates with correctness and flagging failure cases before a full deployment burns budget. A real-LLM evaluation on JudgeBench-style preference tasks is left for future work. The rule is plug-in, and should be a small but useful addition to any multi-agent debate stack.

\section*{Acknowledgments}

Thanks to the open-source maintainers of the Society-of-Minds debate prompts, and to the authors of MMLU, GSM8K, and JudgeBench: the public benchmarks they released made it possible to assess this rule.

\bibliographystyle{IEEEtran}
\bibliography{references}
\end{document}